\documentclass[10pt,letterpaper]{article}

\usepackage{cogsci}
\usepackage{pslatex}
\usepackage{apacite}
\usepackage{graphicx}
\usepackage{color}
\usepackage{natbib}

\title{Blurred Images Lead to Bad Local Minima}
\author{
{\large \bf Gal Katzhendler (gal.katzhendler@mail.huji.ac.il)} \\   Computer Science and Engineering, HUJI
  \AND {\large \bf Daphna Weinshall (daphna@mail.huji.ac.il)} \\ School of Computer Science and Engineering, HUJI
}

\begin{document}

\maketitle

\begin{abstract}
High Initial visual Acuity (HIA) in newborns treated for cataracts, it is argued by \citet{Vogelsang11333}, may cause impairments in configural face analysis. This HIA hypothesis is contrary to the standard explanation by a critical period for learning face processing. The hypothesis is supported by computational experiments with an artificial neural network. In our work we argue that the computational methodology used to evaluate the HIA hypothesis is flawed. It essentially shows that when a classifier is tested with images from different resolutions, the classifier benefits from seeing images from different resolution during its training. We therefore offer a better-fitting methodology; employing the modified methodology - the HIA hypothesis does not hold in simulations using the same artificial neural network model, and the same data. \citet{Vogelsang11333} also show that initial exposure to low resolution images gives rise to larger receptive fields. Our last set of experiments tests the hypothesis that this might be the underlying reason for the observed impairments. Once again, we are unable to find an advantage to training with images of low initial acuity. We therefore conclude that simulations with artificial networks do not support the hypothesis that High Initial visual Acuity is detrimental.

\end{abstract}

\section{Introduction}

In recent years neural networks and deep learning have advanced the state of the art in machine learning and computer vision, where Convolutional Neural Networks (CNNs) in particular have come to dominate visual object recognition. Since its inception the CNN model, inspired by the structure of the visual system, has been used to simulate and investigate the different behaviour and functionality of the visual cortex \citep[e.g.][]{Kaligh2014}, or even predict responses of different parts of the human cortex \citep[e.g.][]{Yamins8619}. With the recent technological advances in this area, and in particular the availability of very powerful and easy to use models, it is now possible to analyze behaviour and functionality more complex than ever before. Vice versa, learning algorithms are continuously inspired by biological and human learning mechanisms \citep{nayebi2017biologically,cruse1998walknet}, in a scientific endeavour where both fields nurture each other.

In accordance, \citet{Vogelsang11333} employed the CNN model when investigating properties of the visual system and its development. They studied a phenomenon whose underlying causes are not well understood, and which concerns children born with bilateral congenital cataract. In one such case, involving a child whose cataract was removed after the age of 4.5 years, the child subsequently regained normal vision but struggled with whole face recognition. The standard explanation for this phenomenon is that facial recognition is an "experience expectant" process, i.e. that there's a critical period early in development that requires exposure to faces in order to develop the necessary facial recognition ability \citep{roder2013sensitive}. 

In their paper, \citet{Vogelsang11333} proposed an alternative explanation. According to their premise, the low initial acuity a newborn experiences due to retinal and cortical immaturities helps to develop larger receptive fields, which are instrumental for the development of integrated spatial vision \citep{witthoft2016reduced}. Thus, according to this hypothesis, the cause of the perceptual deficiency is the High Initial visual Acuity (HIA) that newborns treated for cataracts are exposed to, or more precisely - the lack of initial exposure to low visual acuity. This is henceforth termed the \emph{HIA hypothesis}.

In order to corroborate this proposal and inspect its computational plausibility, \citet{Vogelsang11333} turned to neural networks, and specifically adopted the CNN variant described in \citep{Krizhevsky:2012:ICD:2999134.2999257}. They used different training protocols to simulate the development of the visual system of both children with high initial acuity and normal (low) initial acuity. Low acuity stimuli, experienced normally by newborns \citep[less than 20/600, see][]{dobson1978visual}, was accomplished by blurring the input submitted to the neural networks for training using a Gaussian kernel with $\sigma = 4$. 

Importantly, \citet{Vogelsang11333} evaluated the performance of the different networks, obtained by the different training protocols, using images blurred with a range of different Gaussian kernels. The ecological relevance of blurred images at later stages of development may be attributed to the fact that faces seen from far away take a smaller amount of pixels in the image. When measuring performance based on the integration of accuracy over all blur values, \citet{Vogelsang11333} reported that training with normal low initial acuity outperformed abnormal training with initial high acuity. This result offers a simpler explanation to the clinical observations as compared to the standard explanation, which requires the existence of a "critical period", and should therefore be preferred.


In this paper we challenge this conclusion. Although it is appealing to have an explanation which does not require a "critical period", we argue that the computational study cannot be relied upon to corroborate it, as outlined in the next section. 

\section{Apparent Downside of Low Initial Acuity}

We first focus on the evaluation method by which the "normal" network was found to be beneficial. We note that the performance of the network trained with normal low initial acuity (henceforth denoted \emph{normal network}), as reported in Fig.~3 of \citep{Vogelsang11333}, was in effect \textbf{inferior} to the performance of the network trained with initial high acuity (henceforth denoted \emph{abnormal network}). Only by integrating performance over low resolution images as well, does one see any advantage for the \emph{normal network}. 

But if low resolution images are used for evaluation, which implies that they are ecologically relevant at test time for the mature visual system, then they can surely be used for late training as well. We therefore introduced another control condition, where the network is trained with both low and high resolution images, mixed rather than being introduced consecutively. Surprisingly, this training protocol outperforms the \emph{normal network} over the whole range of blue values. Quite conclusively, a network trained with a mixture of the low and high resolution images outperforms a network trained with the low resolution images first, followed by the high resolution images. Thus, our first conclusion is that the computational study reported in \citep{Vogelsang11333} cannot be used as evidence that initial exposure to low resolution images is beneficial. What it does show is that if a network is tested with both high and low resolution images, then it is beneficial to pre-train it with both high and low resolution images, preferably in a random order.

To further pursue this point, we modified the training protocols of both the \emph{normal} and \emph{abnormal networks}, so that both are exposed to a mix bag of high- and low-resolution images at the second stage of training. Thus modified, the \emph{abnormal network} still outperforms the \emph{normal network} over the whole range of blur values. Since low resolution images are not likely to be ecologically relevant to a mature visual system, we replace all low resolution images in the second stage of training and in the evaluation period by reduced  size images, which better correspond with the rationale for showing low resolution images to a mature visual system - the understanding that when viewed at a distance, objects may appear at low resolution. Even with this modification, the \emph{abnormal network} still outperforms the \emph{normal network} over the whole range of reduced image sizes.

Another interesting observation, reported in \citep{Vogelsang11333}, concerns the size of the receptive fields. Indeed, when inspecting the size of the convolutional filters in the trained artificial networks, \emph{normal networks} are characterized by larger sizes as compared to \emph{abnormal networks}. Is this the property which benefits the recognition of spatially extended objects, like faces? To test this hypothesis, we adopt the feature transfer paradigm commonly used in computational deep learning. We test the beneficial value of the learned filters when used to generalize to other datasets, including images of spatially extended natural scenes. Our results show that in artificial CNN networks, features learned by an \emph{abnormal network} generalize better to other visual tasks such as scene recognition, even though the learned features have smaller receptive fields. Once again, we are unable to find an advantage to training with images of initial low acuity.

The next sections describe these steps in greater detail.

\section{Original Experiments with Additional Control}

We first look at the effect of exposure to different resolutions during training on the final performance of a network. More specifically, we follow five protocols to investigate the effect of showing blurred images from the training data during training, at different schedules. The first four protocols are adapted from \citep{Vogelsang11333}, each involving a total of 500 training epochs:
    \begin{enumerate}
        \item \textbf{Low Resolution to High Resolution (LH)}: a network is trained for 250 epochs with blurred low-resolution images from the training set, followed by 250 training epochs with the original high-resolution images.
        \item \textbf {Only High Resolution (HH)}: a network is trained for 500 epochs with the original images from the training set.
        \item \textbf {High Resolution to Low Resolution (HL)}: a network is trained for 250 epochs with high-resolution images from the training set, followed by 250 epochs of training with blurred low-resolution images from the training set.
        \item \textbf {Only Low Resolution (LL)}: a network is trained for 500 epochs with blurred low-resolution images from the training set.
    \end{enumerate}
We add the following control condition:
    \begin{enumerate}
    \setcounter{enumi}{4}
        \item \textbf {Mixed Low and High Resolutions (Mixed)}: based on the recorded effectiveness of data augmentation in deep learning, we use the blurred images from the training set to augment the training data. Thus a network is trained for 500 epochs with a training set twice as big, keeping the same number of steps-per-epoch as before. In this dataset, each original image from the training set has two instances: high-resolution and low-resolution.
    \end{enumerate}
Recall that protocol~1 attempts to simulate the stimuli that a "normal" newborn experiences (low-resolution to high-resolution) vs. protocol~2 that simulates the stimuli that a newborn with abnormally high initial visual acuity experiences. Our main goal it to compare accuracy in facial recognition between networks trained with these two protocols. The other three protocols are used as control conditions.

Following \citep{Vogelsang11333}, we evaluate the accuracy of neural networks trained with protocols 1-5 over a modified test set, each time blurred by a Gaussian kernel of width $\sigma = 0 \ldots 4$, where $\sigma = 0$ corresponds to the original test set. The rationale behind this evaluation methodology, as explained in the introduction, is that different amounts of Gaussian blur model size variability in images of objects seen from different distances. If this rationale is accepted, then clearly a well rounded visual system should correspond to high accuracy over the whole range of different amounts of blurring. 

    \begin{figure}[ht]
    \begin{center}
    \includegraphics[width=0.95\columnwidth]{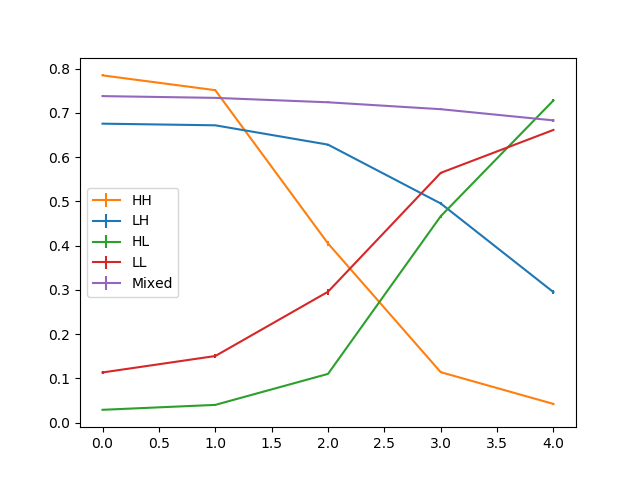}
    \end{center}
    \caption{X-axis: the width of the Gaussian kernel ($\sigma$) applied to each image; Y-axis: accuracy. Error bars show STE with 10 repetitions.} 
    \label{fig:original-results}
    \end{figure}

Results are shown in Fig~\ref{fig:original-results}, in agreement with \citep{Vogelsang11333}. Indeed, if measuring performance over the whole range of Gaussian blurs (e.g., by measuring the area under the curve), it seems that the low-resolution to high-resolution scheme (protocols 1) achieves the best performance when comparing the first 4 protocols only. However, note that the 5th protocol, not included in the study of \citep{Vogelsang11333}, performs even better when employing this  integrative measure. What is the significance of this result?

If we accept the premise outlined above that a visual system is likely to encounter images modeled by a variety of blurring levels, we should likewise modify the way we model the visual experiences of both normal and abnormal visual systems. Specifically, if blurred images are part of the normal visual experience even in later stages of life, after the retina has matured, then the experience of an abnormal baby, modeled by protocol 2, should be changed to protocol 5. We therefore conclude that the results shown in Fig~\ref{fig:original-results}, replicating the results reported in \citep{Vogelsang11333}, in effect demonstrate that the protocol which best models the abnormal baby achieves best performance in the proposed integrative measure of success. 

We note that the experience of a normal baby, modeled by protocol 1, should also be changed to reflect this issue. This is discussed in the next section.


\section{Reconsidering the Original Methodology} 
\label{reconsidering}

The rationale behind the test paradigm used in \citep{Vogelsang11333} and replicated above is based on the premise that classification accuracy at different levels of blurring is ecologically relevant. If this premise is correct, it implies the validity of a stronger premise: humans are continuously exposed to low resolution images of objects even after their visual system matures, possibly because objects are observed at different distances. This final premise implies that the training of a mature visual system should involve both high and low resolution images as in protocol 5 (mixed). 

\begin{figure}[ht]
    \begin{center}
    \includegraphics[width=0.95\columnwidth]{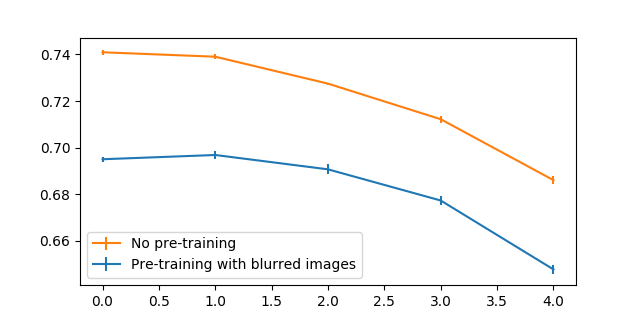}
    \end{center}
    \caption{X-axis: $\sigma$ of Gaussian kernel; Y-axis: accuracy. Error bars show STE with 3 repetitions.} 
    \label{fig:new-methodology}
    \end{figure}

In our next experimental setup we compare the different training protocols modified to reflect this more consistent experimental methodology. Thus we compare two networks:
\begin{enumerate}
    \item \textbf{Pre-training with blurred images}: a network is trained for 250 epochs with blurred low-resolution images ($\sigma = 4$ as in protocol 1), followed by 500 training epochs with a mix of high resolution and low resolution images.
    \item \textbf{No pre-training}: a network is trained for 500 epochs with a mix of high resolution and low resolution images.
\end{enumerate}
Results are shown in Fig.~\ref{fig:new-methodology}.

The results when using the modified methodology show a robust and significant advantage for the network that was not pre-trained with blurred images (protocol~2). This result is very interesting given the fact that the pre-trained network is trained with a modified protocol 1 that involves 250 epochs more than protocol 2, and where the last 500 epochs are exactly the same as the whole of protocol 2. Thus, it seems that pre-training has a detrimental effect on the final performance of this network, possibly by biasing the random weight initialization in a non-optimal way.

\section{Blurred, or Reduced-Size Images?}

Going back to the original results of \citep{Vogelsang11333} as replicated in Fig~\ref{fig:original-results}, we recall that the rationale for evaluating performance on blurred images is the notion that humans need to recognize low resolution images of objects when these objects are seen at a distance. This suggests that, rather than using low resolution images in the train or test set, it is more ecological in a mature system to use reduced size images of objects. This will provide a better model for the the affect of distance on resolution in terms of the frequency coding of the image. 


    \begin{figure}[ht]
    \begin{center}
    \includegraphics[width=8cm]{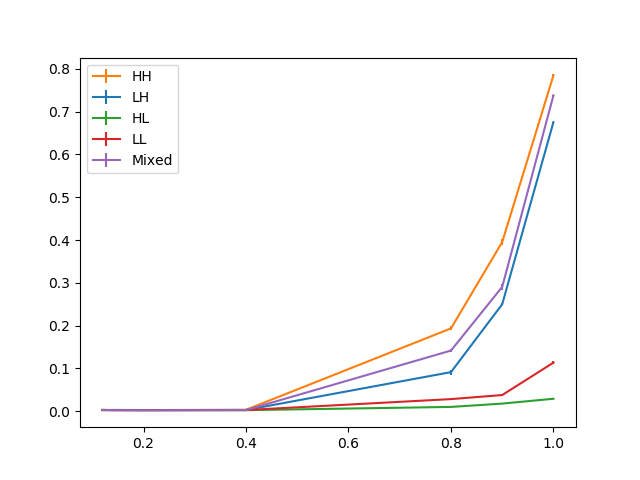}
    \end{center}
    \caption{X-axis: shrinking scaling factor in each dimension; Y-axis: accuracy. Error bars show STE after 3 repetitions.}
    \label{fig:over_dists}
    \end{figure}

We therefore repeat the same empirical investigation as described above, but in each case during the evlauation stage - instead of using an image blurred by some Gaussian kernel parameterized by $\sigma$, we shrink the corresponding image to achieve the same effective resolution, and locate it in the center of a regular size image while padding the rest of the area with zeros. Accordingly, the linear shrinking scale factor (per dimension) is chosen in the range $[1, 0.9, 0.8, 0.4, 0.2, 0.14, 0.12]$, in order to generate images with approximately similar resolution levels as generated by the relevant range of Gaussian kernels used in the original experiments ($[1, 0.4, 0.2, 0.14, 0.12]$). 

Fig.~\ref{fig:over_dists} shows the results when using the networks trained with protocols 1-5 as described above. This time the advantage of protocol 2 is even clearer: protocol 2 (the \emph{abnormal network}) outperforms protocol 1 (the \emph{normal network}) over the whole range of shrinking scaling factors. 

\subsection{Methodology Reconsidered}

We return to the modified methodology, as described in the previous section, and repeat the experiments where low resolution is achieved by reducing the size of the images:
\begin{enumerate}
    \item \textbf{Low Initial Acuity}: a network is trained for 250 epochs with blurred images, followed by training for 250 epochs with the regular training set where we shrink and center the image with probability 0.5.
    \item \textbf{High Initial Acuity}: a network is trained for 500 epochs with the regular training set, where we shrink and center the image with probability 0.5.
\end{enumerate}
These training protocols reflect our perception that images of distant objects are not necessarily equivalent to images of blurred objects. Protocol 1 now simulates the condition of a "normal" newborn having initial low visual acuity, where the rest of the training reflects the vision of a "normal" human with developed eye-sight, exposed to objects both near and far.

    \begin{figure}[ht]
    \begin{center}
    \includegraphics[width=0.95\columnwidth]{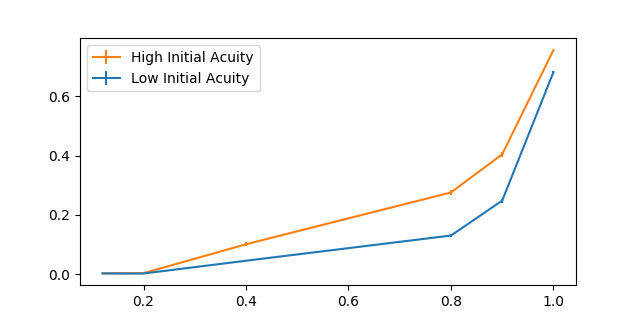}
    \end{center}
    \caption{X-axis: shrinking factor along each dimension; Y-axis: accuracy. Error bars show STE for 3 repetitions.} 
    \label{fig:shrunk_over_dists}
    \end{figure}

Results are shown in Fig.~\ref{fig:shrunk_over_dists}. Once again, protocol 2 performs as well or better than protocol 1 for any amount of shrinking. Like before, it seems that pre-training with blurred images degrades, or at least doesn't benefit, the accuracy of the network in ecologically relevant tasks.

\section{Spatial Integration and Generalization}

When inspecting the details of the corresponding networks trained with protocols 1 and 2, we see that protocol 1 promotes the buildup of filters (or convolutions) with wider receptive fields. \citep{Vogelsang11333} reported this observation, and hypothesized that these larger receptive fields provide the network with extended spatial integration abilities, which may facilitate better performance in such tasks as whole face recognition. In other words, this hypothesis predicts that networks trained with protocol 1 should demonstrate better generalization, or better transfer ability, to tasks requiring whole image processing, when compared to networks trained with protocol 2. 

To study this hypothesis, we propose to investigate the efficacy of the learned features in transfer learning to new visual object recognition tasks, where  whole image integration is likely to be beneficial. To this end we adopt a common feature transfer paradigm used in deep learning \citep{sharif2014cnn}, where features learned by a network pre-trained in one task are used to accelerate and enhance the learning in another task. Specifically, the activation in one hidden layer\footnote{Phrased in this manner, each hidden layer of the pre-trained network gives rise to a different representation.} in the pre-trained network is used to generate the transfer representation for all the images in the new task, both in the train and test sets. This representation is subsequently used while training a simpler classifier (here we use the linear SVM) to perform the new classification task. 

\subsection{Feature transfer in the same domain}

    \begin{figure}[ht]
    \begin{center}
    \includegraphics[width=8cm]{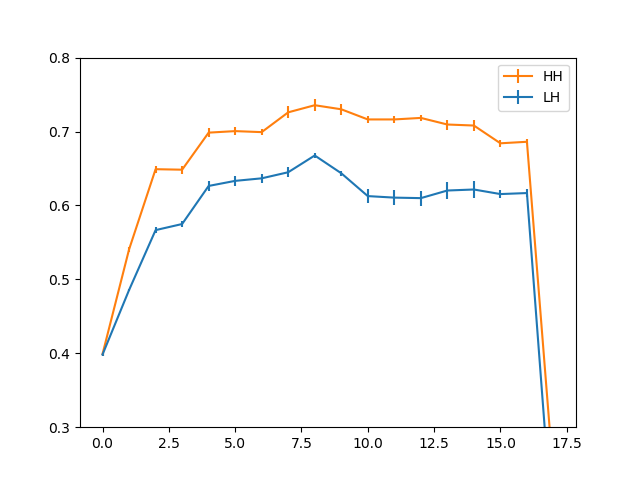}
    \end{center}
    \caption{X-axis: layer from which the transfer representations are extracted; Y-axis: accuracy in the recognition of unseen faces from the same data set.  Error bars show STE with 4 iterations. Blurring kernel has $\sigma = 4$.} 
    \label{fig:tansfer-faces}
    \end{figure}

We first use the feature transfer paradigm for the recognition of unseen faces from the original data set ($n<100$), assuming that faces require extended spatial integration and therefore facial images may show the highest benefit. Results are shown in Fig.~\ref{fig:tansfer-faces}, demonstrating a clear edge for protocol 2 over protocol 1, contrary to the prediction of the tested hypothesis. This implies that features based on extended receptive fields do not always benefit transfer in the recognition of facial images. In fact, the less extended features learned by protocol 2 seem better suited for the recognition of the new faces.

    \begin{figure}[ht]
    \begin{center}
    \includegraphics[width=0.8\columnwidth]{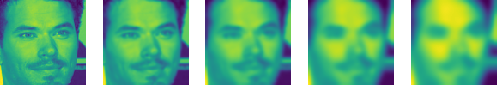}\\
    \includegraphics[width=0.8\columnwidth]{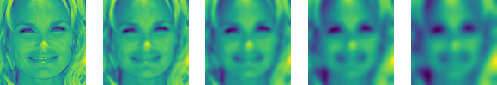}\\
    \includegraphics[width=0.8\columnwidth]{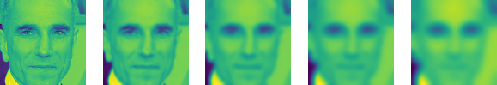} \begin{tabular*}{0.8\columnwidth}{ccccc}
    $\sigma=0$\hspace{0.02\columnwidth}  &    $\sigma=1$\hspace{0.02\columnwidth}  &    $\sigma=2$\hspace{0.02\columnwidth}  &    $\sigma=3$\hspace{0.02\columnwidth}  &    $\sigma=4$ \\ 
    \end{tabular*}
    \end{center}
    \caption{Examples of facial images blurred by a Gaussian filter with different kernels.} 
    \label{fig:blurred-images}
    \end{figure}

\begin{figure}[ht]
    \begin{center}
    \includegraphics[width=0.95\columnwidth]{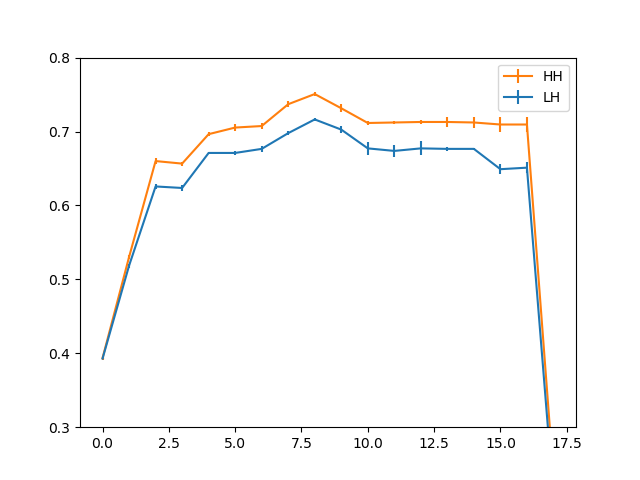}
    \end{center}
    \caption{X-axis: layer from which the representations are extracted; Y-axis: accuracy. Error bars show STE with 4 repetitions. Blurring kernel has $\sigma = 2$.}
    \label{faces_sig_2}
    \end{figure}

We note that $\sigma = 4$ corresponds to a very significant blurring that makes faces hardly distinguishable to human observers (see Fig.~\ref{fig:blurred-images}), and thus may generate features that are too spatially extended to be useful. We therefore repeat the same experiment with a range of different kernels $\sigma = 1, \ldots,4$, where in each training session the set of blurred images is generated by a different Gaussian kernel (see Fig.~\ref{faces_sig_2} for $\sigma = 2$). The results remain qualitatively the same, while showing a decline in the extent to which protocol 2 outperforms protocol 1 as $\sigma$ decreases.

These findings seem to contradict the hypothesis of \citep{Vogelsang11333} that early exposure to blurred images contributes to the acquisition of extended spatial vision abilities, at least in artificial neural networks. The findings suggest that artificial neural networks learn the appropriate spatial extension of the most useful features when exposed to high resolution images at the onset of learning.

\subsection{Feature transfer to other domains}

An important benefit of deep learning is the utility of the features to accelerate the learning of similar problems in other domains, such as the recognition of different types of objects. We now investigate if and how the learned features in the different protocols benefit transfer learning into new types of data sets by using the same feature transfer paradigm. Specifically, we use the following datasets, which have been chosen from domains where extended spatial integration would appear to offer benefits:
\begin{itemize}
    \item \textbf{SCENE67} MIT's Indoor Scene Recognition data set \citep{quattoni2009recognizing}, a challenging data set of 67 different indoor scenes, which requires both spatial features and local features to classify.
    \item \textbf{flowers102} Oxford's 102 category Flowers data set \citep{nilsback2008automated} consisting of common flowers seen in the UK.
\end{itemize}

    \begin{figure}[ht]
    \begin{center}
    \includegraphics[width=0.95\columnwidth]{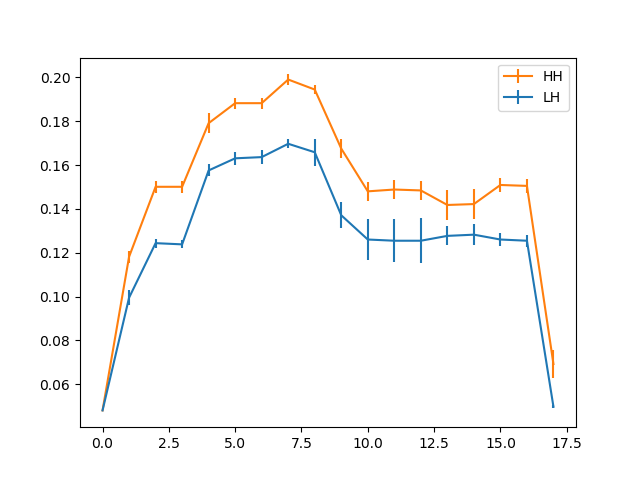}
    \end{center}
    \caption{X-axis: Layer from which the representations are extracted, Y-axis: accuracy. Error bars show STE after 4 repetitions.} 
    \label{fig:transfer-other}
    \end{figure}
    
Results are shown in Fig.~\ref{fig:transfer-other}, demonstrating some amount of improvement based on transfer from faces to SCENE67, while little improvement is seen with transfer to flowers102. Still, features based on training with protocol 1 transfer less well than features based on training with protocol 2, showing once more that the more spatially extended features are not beneficial in this generalization challenge.

\section{Methods}

Following are the details of our different implementations:

\subsubsection{Architecture.}
In order to replicate the architecture used in \citep{Vogelsang11333}, we used a variant of AlexNet \citep{Krizhevsky:2012:ICD:2999134.2999257}, consisting of: (i) (conv + max pooling + local response normalization)$\times 2$. (ii) (conv)$\times 3$. (iii) max pooling + local response normalization. (iv) (fully connected + dropout(0.5))$\times 2$. (v) fully connected with softmax activation. Following \citep{Vogelsang11333}, we adjusted the input of the network to be $100 \times 100$, and changed the shape of the filters in the first convolution layer to be $22\times 22$. All of the convolutional layers had ReLU activation functions. All but the last fully connected layers had tanh activation function, while the last layer had softmax activation function. The networks were trained with a learning rate of 0.001 and momentum optimizer with constant 0.9. The batch size while training was 128. 

\subsubsection{Data.}
Similarly to \citep{Vogelsang11333}, the data set used in the experiments was Facescrub \citep{facescrub}, a data set with 67596 face-cropped color images belonging to 530 different face identities. We rejected any identity with less than 100 images, ending up with 438 identities. We also discarded extra images for identities with above 100 images, so over all we had 43800 images with a 90/10 train/test split. All images were converted to grayscale, zero centered to their overall mean of luminance and scaled by the overall standard deviation. During training, the images were randomly flipped horizontally and rotated by an angle of up to $25^{\circ}$. Gaussian blur was applied where indicated with $\sigma = 0\ldots 4$.

\section{Summary and Discussion}

The body of experiments and results described above shows quite consistently that pre-training neural networks with blurred images, before exposing them to high resolution images, does not provide benefit for tasks requiring extended spatial vision, and specifically face recognition. In fact, the inverse is true: our result support the hypothesis that this type of training regime is harmful in such tasks. This computational evidence contradicts the basic premise of the HIA hypothesis, which predicts that high initial visual acuity harms extended spatial vision in humans. This does not necessarily mean that hypothesis is false in humans, only that it is not corroborated by this kind of computational evidence. Further investigation may involve new biological experiments and different computational models, perhaps models that simulate the biological mechanisms of the human visual system with greater fidelity.

For future work, the new evaluation tools discussed above may provide ways to evaluate the amount of extended spatial features acquired by neural networks during training. This can help estimate the effect of Low Initial Acuity on future computational models. 

\bibliographystyle{apacite}

\setlength{\bibleftmargin}{.125in}
\setlength{\bibindent}{-\bibleftmargin}

\bibliography{cogsci_template}

\begin{thebibliography}{}

\bibitem [\protect \citeauthoryear {%
Cruse%
, Kindermann%
, Schumm%
, Dean%
\BCBL {}\ \BBA {} Schmitz%
}{%
Cruse%
\ \protect \BOthers {.}}{%
{\protect \APACyear {1998}}%
}]{%
cruse1998walknet}
\APACinsertmetastar {%
cruse1998walknet}%
\begin{APACrefauthors}%
Cruse, H.%
, Kindermann, T.%
, Schumm, M.%
, Dean, J.%
\BCBL {}\ \BBA {} Schmitz, J.%
\end{APACrefauthors}%
\unskip\
\newblock
\APACrefYearMonthDay{1998}{}{}.
\newblock
{\BBOQ}\APACrefatitle {Walknet—a biologically inspired network to control
  six-legged walking} {Walknet—a biologically inspired network to control
  six-legged walking}.{\BBCQ}
\newblock
\APACjournalVolNumPages{Neural networks}{11}{7-8}{1435--1447}.
\PrintBackRefs{\CurrentBib}

\bibitem [\protect \citeauthoryear {%
Dobson%
\ \BBA {} Teller%
}{%
Dobson%
\ \BBA {} Teller%
}{%
{\protect \APACyear {1978}}%
}]{%
dobson1978visual}
\APACinsertmetastar {%
dobson1978visual}%
\begin{APACrefauthors}%
Dobson, V.%
\BCBT {}\ \BBA {} Teller, D\BPBI Y.%
\end{APACrefauthors}%
\unskip\
\newblock
\APACrefYearMonthDay{1978}{}{}.
\newblock
{\BBOQ}\APACrefatitle {Visual acuity in human infants: a review and comparison
  of behavioral and electrophysiological studies} {Visual acuity in human
  infants: a review and comparison of behavioral and electrophysiological
  studies}.{\BBCQ}
\newblock
\APACjournalVolNumPages{Vision research}{18}{11}{1469--1483}.
\PrintBackRefs{\CurrentBib}

\bibitem [\protect \citeauthoryear {%
Khaligh-Razavi%
\ \BBA {} Kriegeskorte%
}{%
Khaligh-Razavi%
\ \BBA {} Kriegeskorte%
}{%
{\protect \APACyear {2014}}%
}]{%
Kaligh2014}
\APACinsertmetastar {%
Kaligh2014}%
\begin{APACrefauthors}%
Khaligh-Razavi, S\BHBI M.%
\BCBT {}\ \BBA {} Kriegeskorte, N.%
\end{APACrefauthors}%
\unskip\
\newblock
\APACrefYearMonthDay{2014}{}{}.
\newblock
{\BBOQ}\APACrefatitle {Deep supervised, but not unsupervised, models may
  explain IT cortical representation} {Deep supervised, but not unsupervised,
  models may explain it cortical representation}.{\BBCQ}
\newblock
\APACjournalVolNumPages{PLoS computational biology}{10}{11}{e1003915}.
\PrintBackRefs{\CurrentBib}

\bibitem [\protect \citeauthoryear {%
Krizhevsky%
, Sutskever%
\BCBL {}\ \BBA {} Hinton%
}{%
Krizhevsky%
\ \protect \BOthers {.}}{%
{\protect \APACyear {2012}}%
}]{%
Krizhevsky:2012:ICD:2999134.2999257}
\APACinsertmetastar {%
Krizhevsky:2012:ICD:2999134.2999257}%
\begin{APACrefauthors}%
Krizhevsky, A.%
, Sutskever, I.%
\BCBL {}\ \BBA {} Hinton, G\BPBI E.%
\end{APACrefauthors}%
\unskip\
\newblock
\APACrefYearMonthDay{2012}{}{}.
\newblock
{\BBOQ}\APACrefatitle {ImageNet Classification with Deep Convolutional Neural
  Networks} {Imagenet classification with deep convolutional neural
  networks}.{\BBCQ}
\newblock
\BIn{} \APACrefbtitle {Proceedings of {NIPS}} {Proceedings of {NIPS}}\ (\BPGS\
  1097--1105).
\PrintBackRefs{\CurrentBib}

\bibitem [\protect \citeauthoryear {%
Nayebi%
\ \BBA {} Ganguli%
}{%
Nayebi%
\ \BBA {} Ganguli%
}{%
{\protect \APACyear {2017}}%
}]{%
nayebi2017biologically}
\APACinsertmetastar {%
nayebi2017biologically}%
\begin{APACrefauthors}%
Nayebi, A.%
\BCBT {}\ \BBA {} Ganguli, S.%
\end{APACrefauthors}%
\unskip\
\newblock
\APACrefYearMonthDay{2017}{}{}.
\newblock
{\BBOQ}\APACrefatitle {Biologically inspired protection of deep networks from
  adversarial attacks} {Biologically inspired protection of deep networks from
  adversarial attacks}.{\BBCQ}
\newblock
\APACjournalVolNumPages{arXiv preprint arXiv:1703.09202}{}{}{}.
\PrintBackRefs{\CurrentBib}

\bibitem [\protect \citeauthoryear {%
Ng%
\ \BBA {} Winkler%
}{%
Ng%
\ \BBA {} Winkler%
}{%
{\protect \APACyear {2014}}%
}]{%
facescrub}
\APACinsertmetastar {%
facescrub}%
\begin{APACrefauthors}%
Ng, H\BHBI W.%
\BCBT {}\ \BBA {} Winkler, S.%
\end{APACrefauthors}%
\unskip\
\newblock
\APACrefYearMonthDay{2014}{}{}.
\newblock
{\BBOQ}\APACrefatitle {A data-driven approach to cleaning large face datasets}
  {A data-driven approach to cleaning large face datasets}.{\BBCQ}
\newblock
\BIn{} \APACrefbtitle {Proceedings of {IEEE ICIP}} {Proceedings of {IEEE
  ICIP}}\ (\BPG~343-347).
\PrintBackRefs{\CurrentBib}

\bibitem [\protect \citeauthoryear {%
Nilsback%
\ \BBA {} Zisserman%
}{%
Nilsback%
\ \BBA {} Zisserman%
}{%
{\protect \APACyear {2008}}%
}]{%
nilsback2008automated}
\APACinsertmetastar {%
nilsback2008automated}%
\begin{APACrefauthors}%
Nilsback, M\BHBI E.%
\BCBT {}\ \BBA {} Zisserman, A.%
\end{APACrefauthors}%
\unskip\
\newblock
\APACrefYearMonthDay{2008}{}{}.
\newblock
{\BBOQ}\APACrefatitle {Automated flower classification over a large number of
  classes} {Automated flower classification over a large number of
  classes}.{\BBCQ}
\newblock
\BIn{} \APACrefbtitle {Proceedings of {IEEE ICVGIP}} {Proceedings of {IEEE
  ICVGIP}}\ (\BPGS\ 722--729).
\PrintBackRefs{\CurrentBib}

\bibitem [\protect \citeauthoryear {%
Quattoni%
\ \BBA {} Torralba%
}{%
Quattoni%
\ \BBA {} Torralba%
}{%
{\protect \APACyear {2009}}%
}]{%
quattoni2009recognizing}
\APACinsertmetastar {%
quattoni2009recognizing}%
\begin{APACrefauthors}%
Quattoni, A.%
\BCBT {}\ \BBA {} Torralba, A.%
\end{APACrefauthors}%
\unskip\
\newblock
\APACrefYearMonthDay{2009}{}{}.
\newblock
{\BBOQ}\APACrefatitle {Recognizing indoor scenes} {Recognizing indoor
  scenes}.{\BBCQ}
\newblock
\BIn{} \APACrefbtitle {Proceedings of {IEEE CVPR}} {Proceedings of {IEEE
  CVPR}}\ (\BPGS\ 413--420).
\PrintBackRefs{\CurrentBib}

\bibitem [\protect \citeauthoryear {%
R{\"o}der%
, Ley%
, Shenoy%
, Kekunnaya%
\BCBL {}\ \BBA {} Bottari%
}{%
R{\"o}der%
\ \protect \BOthers {.}}{%
{\protect \APACyear {2013}}%
}]{%
roder2013sensitive}
\APACinsertmetastar {%
roder2013sensitive}%
\begin{APACrefauthors}%
R{\"o}der, B.%
, Ley, P.%
, Shenoy, B\BPBI H.%
, Kekunnaya, R.%
\BCBL {}\ \BBA {} Bottari, D.%
\end{APACrefauthors}%
\unskip\
\newblock
\APACrefYearMonthDay{2013}{}{}.
\newblock
{\BBOQ}\APACrefatitle {Sensitive periods for the functional specialization of
  the neural system for human face processing} {Sensitive periods for the
  functional specialization of the neural system for human face
  processing}.{\BBCQ}
\newblock
\APACjournalVolNumPages{Proceedings of the National Academy of
  Sciences}{}{}{201309963}.
\PrintBackRefs{\CurrentBib}

\bibitem [\protect \citeauthoryear {%
Sharif~Razavian%
, Azizpour%
, Sullivan%
\BCBL {}\ \BBA {} Carlsson%
}{%
Sharif~Razavian%
\ \protect \BOthers {.}}{%
{\protect \APACyear {2014}}%
}]{%
sharif2014cnn}
\APACinsertmetastar {%
sharif2014cnn}%
\begin{APACrefauthors}%
Sharif~Razavian, A.%
, Azizpour, H.%
, Sullivan, J.%
\BCBL {}\ \BBA {} Carlsson, S.%
\end{APACrefauthors}%
\unskip\
\newblock
\APACrefYearMonthDay{2014}{}{}.
\newblock
{\BBOQ}\APACrefatitle {CNN features off-the-shelf: an astounding baseline for
  recognition} {Cnn features off-the-shelf: an astounding baseline for
  recognition}.{\BBCQ}
\newblock
\BIn{} \APACrefbtitle {Proceedings of {IEEE CVPR}} {Proceedings of {IEEE
  CVPR}}\ (\BPGS\ 806--813).
\PrintBackRefs{\CurrentBib}

\bibitem [\protect \citeauthoryear {%
Vogelsang%
\ \protect \BOthers {.}}{%
Vogelsang%
\ \protect \BOthers {.}}{%
{\protect \APACyear {2018}}%
}]{%
Vogelsang11333}
\APACinsertmetastar {%
Vogelsang11333}%
\begin{APACrefauthors}%
Vogelsang, L.%
, Gilad-Gutnick, S.%
, Ehrenberg, E.%
, Yonas, A.%
, Diamond, S.%
, Held, R.%
\BCBL {}\ \BBA {} Sinha, P.%
\end{APACrefauthors}%
\unskip\
\newblock
\APACrefYearMonthDay{2018}{}{}.
\newblock
{\BBOQ}\APACrefatitle {Potential downside of high initial visual acuity}
  {Potential downside of high initial visual acuity}.{\BBCQ}
\newblock
\APACjournalVolNumPages{Proceedings of the National Academy of
  Sciences}{115}{44}{11333--11338}.
\PrintBackRefs{\CurrentBib}

\bibitem [\protect \citeauthoryear {%
Witthoft%
\ \protect \BOthers {.}}{%
Witthoft%
\ \protect \BOthers {.}}{%
{\protect \APACyear {2016}}%
}]{%
witthoft2016reduced}
\APACinsertmetastar {%
witthoft2016reduced}%
\begin{APACrefauthors}%
Witthoft, N.%
, Poltoratski, S.%
, Nguyen, M.%
, Golarai, G.%
, Liberman, A.%
, LaRocque, K.%
\BCBL {}\ \BBA {} Grill-Spector, K.%
\end{APACrefauthors}%
\unskip\
\newblock
\APACrefYearMonthDay{2016}{}{}.
\newblock
\APACrefbtitle {Reduced spatial integration in the ventral visual cortex
  underlies face recognition deficits in developmental prosopagnosia (No.
  biorxiv; 051102v1).} {Reduced spatial integration in the ventral visual
  cortex underlies face recognition deficits in developmental prosopagnosia
  (no. biorxiv; 051102v1).}
\PrintBackRefs{\CurrentBib}

\bibitem [\protect \citeauthoryear {%
Yamins%
\ \protect \BOthers {.}}{%
Yamins%
\ \protect \BOthers {.}}{%
{\protect \APACyear {2014}}%
}]{%
Yamins8619}
\APACinsertmetastar {%
Yamins8619}%
\begin{APACrefauthors}%
Yamins, D\BPBI L\BPBI K.%
, Hong, H.%
, Cadieu, C\BPBI F.%
, Solomon, E\BPBI A.%
, Seibert, D.%
\BCBL {}\ \BBA {} DiCarlo, J\BPBI J.%
\end{APACrefauthors}%
\unskip\
\newblock
\APACrefYearMonthDay{2014}{}{}.
\newblock
{\BBOQ}\APACrefatitle {Performance-optimized hierarchical models predict neural
  responses in higher visual cortex} {Performance-optimized hierarchical models
  predict neural responses in higher visual cortex}.{\BBCQ}
\newblock
\APACjournalVolNumPages{Proceedings of the National Academy of
  Sciences}{111}{23}{8619--8624}.
\PrintBackRefs{\CurrentBib}

\end{thebibliography}

\end{document}